\documentclass[runningheads]{lib/llncs}
\usepackage{graphicx}
\usepackage{xcolor}
\usepackage{amsmath} % define this before the line numbering.
\usepackage{booktabs} % To thicken table lines

\begin{document}
% \linenumbers
\pagestyle{headings}
\mainmatter
\def\ACCV20SubNumber{1036}  % Insert your submission number here

\title{Trainable Structure Tensors for Autonomous Baggage Threat Detection Under Extreme Occlusion} % Replace with your title

% % CAMERA READY SUBMISSION
 \titlerunning{Trainable Structure Tensors}
% % If the paper title is too long for the running head, you can set
% % an abbreviated paper title here
% %
 \author{Taimur Hassan\inst{1} \and
 Samet Ak\c{c}ay\inst{2}\and
 Mohammed Bennamoun\inst{3} \and
 Salman Khan\inst{4} \and
 Naoufel Werghi \inst{1}
 }
% %
 \authorrunning{T. Hassan et al.}
 \institute{Center for Cyber-Physical Systems (C2PS), Khalifa University, UAE \\ \email{\{taimur.hassan, naoufel.werghi\}@ku.ac.ae} \and
 Department of Computer Science, Durham University, UK \\ 
 \email{samet.akcay@durham.ac.uk} \and
 Department of Computer Science and Software Engineering, The University of Western Australia, Australia\\ 
 \email{mohammed.bennamoun@uwa.edu.au} \and
 Department of Computer Vision, Mohamed Bin Zayed University of Artificial Intelligence, UAE\\
 \email{salman.khan@mbzuai.ac.ae}
 }
 
\maketitle
\begin{abstract}
Detecting baggage threats is one of the most difficult tasks, even for expert officers. Many researchers have developed computer-aided screening systems to recognize these threats from the baggage X-ray scans. However, all of these frameworks are limited in identifying the contraband items under extreme occlusion. This paper presents a novel instance segmentation framework that utilizes trainable structure tensors to highlight the contours of the occluded and cluttered contraband items (by scanning multiple predominant orientations), while simultaneously suppressing the irrelevant baggage content. The proposed framework has been extensively tested on four publicly available X-ray datasets where it outperforms the state-of-the-art frameworks in terms of mean average precision scores. Furthermore, to the best of our knowledge, it is the only framework that has been validated on combined grayscale and colored scans obtained from four different types of X-ray scanners.

\end{abstract}

\section{Introduction}
Detecting threats concealed within the baggage has gained the utmost attention of aviation staff throughout the world. While X-ray imagery provides a thorough insight into the baggage content, manually screening them is a very cumbersome task, requiring constant attention of the human observer. To cater this, many researchers have developed autonomous frameworks for recognizing baggage threats from the security X-ray scans. At first, these frameworks employed conventional machine learning to identify contraband items. However, due to the subjectiveness in their feature sets, they were confined to small-scale datasets and limited experimental settings. More recently, deep learning has boosted the performance of baggage threat detection frameworks. In this paper, we discuss the pioneer works for detecting baggage threats, and for a detailed survey, we refer the readers to the work of \cite{ackay2020}.

\subsection{Conventional Machine Learning Methods}
Initial methods developed for screening prohibited items used classification \cite{turcsany2013improving}, detection \cite{bastan2015} and segmentation \cite{heitz2010} strategies. The classification schemes are driven through hand-engineered features \cite{zhang2014} \cite{jaccard2014automated} and key-point descriptors such as SURF \cite{bastan2011}, FAST-SURF \cite{kundegorski2016} and SIFT \cite{mery2016} \cite{zhang2014} in conjunction with Support Vector Machines (SVM) \cite{bastan2011} \cite{turcsany2013improving} \cite{kundegorski2016}, Random Forest \cite{jaccard2014automated} and K-Nearest Neighbor (K-NN) \cite{riffo2015automated} models for recognizing baggage threats. Contrary to this, researchers have also proposed object detection schemes employing fused SPIN and SIFT descriptors derived from multi-view X-ray imagery \cite{bastan2015}. Moreover, 3D feature points driven through the structure from motion have also been utilized in recognizing the potential baggage threats \cite{mery2016}. Apart from this, the segmentation schemes utilizing region-growing and SURF features along with the Neighbor Distance achieved good performance for recognizing prohibited baggage items \cite{heitz2010}.  

\subsection{Deep Learning Methods}
The initial wave of deep learning methods employed transfer learning \cite{akcay2018using} \cite{jaccard2017} for recognizing the baggage threats followed by object detection \cite{liu2018detection} \cite{Xu2018} \cite{miao2019sixray} \cite{gaus2019evaluating} \cite{hassan2019}, and segmentation strategies \cite{gaus2019evaluation} \cite{an2019}. Recently, researchers used anomaly detection \cite{akcay2018ganomaly} \cite{samet2019} as a means to handle data scarcity while recognizing potential baggage threats. Moreover, attention modules \cite{Xu2018} and maximum likelihood estimations \cite{griffin2019} have also been explored for detecting the contraband items. Apart from this, Miao et al. \cite{miao2019sixray} exploited the imbalanced and extremely cluttered nature of the baggage threats by introducing a large-scale dataset dubbed Security Inspection X-ray (SIXray) \cite{miao2019sixray}. They also proposed a class-balanced hierarchical framework (CHR) to recognize the contraband items from the highly complex scans of the SIXray \cite{miao2019sixray} dataset. Furthermore, Hassan et al. developed a Cascaded Structure Tensor (CST) framework that alleviates contours of the contraband items to generate object proposals which are classified through the ResNet-50 \cite{he2016deep} model. CST is validated on publicly available GRIMA X-ray Database (GDXray) \cite{mery2015gdxray} and SIXray \cite{miao2019sixray} datasets. Moreover, Wei et al. \cite{opixray} developed De-occlusion Attention Module (DOAM), a plug and play module that can be paired with object detectors to recognize and localize the occluded contraband items from the baggage X-ray scans. DOAM has been rigorously tested on a publicly available Occluded Prohibited Items X-ray (OPIXray) dataset introduced in \cite{opixray}. 

\begin{figure}[t]
    \centering
    \includegraphics[width=1\linewidth]{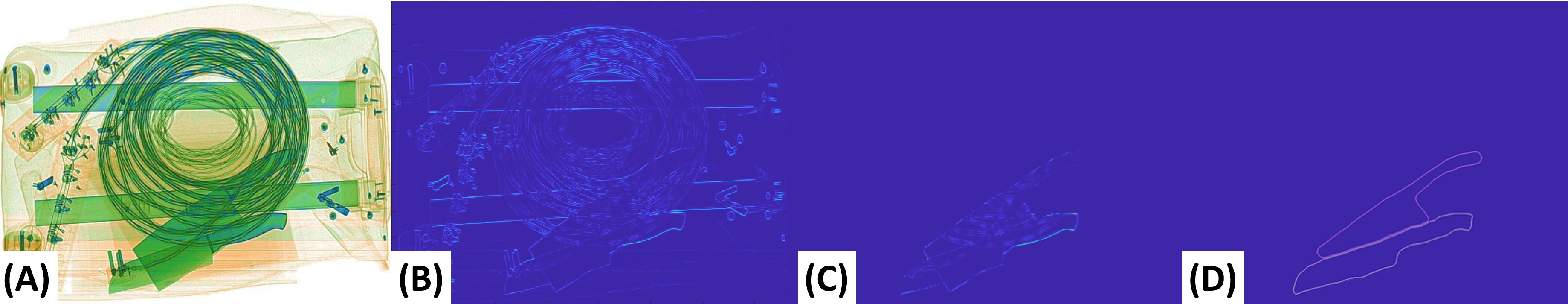}
    \caption{(A) Baggage X-ray scan from the SIXray \cite{miao2019sixray} dataset, (B) best tensor representation obtained through the modified structure tensor approach \cite{hassan2019}, (C) suppression of the irrelevant baggage content through proposed trainable structure tensors, (D) predicted boundaries of the contraband items.}
    \label{fig:fig1}
\end{figure}

\noindent Detecting occluded and extremely cluttered contraband items from the baggage X-ray scans is a very challenging task \cite{akcay2018using} \cite{gaus2019evaluation}. Towards this end, frameworks such as CHR \cite{miao2019sixray}, CST \cite{hassan2019}, and DOAM \cite{opixray} possess the capability to identify occluded baggage items. However, these frameworks are either tested on a single dataset \cite{miao2019sixray} \cite{opixray} or cannot be generalized for the multiple scanner specifications due to exhaustive parametric tuning \cite{hassan2019}. 
To cater these limitations, this paper presents a single-staged instance segmentation framework that leverages proposed trainable structure tensors to recognize the contours of the prohibited items while suppressing the irrelevant baggage content as shown in Fig. \ref{fig:fig1}. To summarize, the main features of the paper are:

\begin{itemize}
    \item A novel trainable structure tensors scheme to highlight transitional patterns of the desired objects by analyzing predominant orientations of the associated image gradients.
    \item A single-staged instance segmentation framework capable of recognizing extremely cluttered and occluded contraband items from scans acquired through diverse ranging scanner specifications.
    \item A rigorous validation of the proposed framework on four publicly available X-ray datasets for recognizing baggage threats under extreme occlusion.
\end{itemize}

\noindent Rest of the paper is organized as follows: Section \ref{sec:proposed} describe the proposed framework in detail, Section \ref{sec:exp} presents the experimental setup along with the datasets description, the implementation details and the evaluation metrics. Section \ref{sec:results} discuss the experimental results and Section \ref{sec:conclusion} concludes the paper.

\section{Proposed Framework} \label{sec:proposed}
The block diagram of the proposed framework is shown in Fig. \ref{fig:fig2}. First of all, we compute the best tensor representation of the input scan through the structure tensor module. Afterward, we pass it through the multi-class encoder-decoder backbone that only retains the transitional patterns of the threatening items while suppressing the rest of the baggage content. The extracted contours are post-processed and then utilized in generating the bounding boxes and masks of the corresponding suspicious items. The detailed description of each module is presented below:

\begin{figure}[htb]
    \centering
    \includegraphics[width=0.95\linewidth]{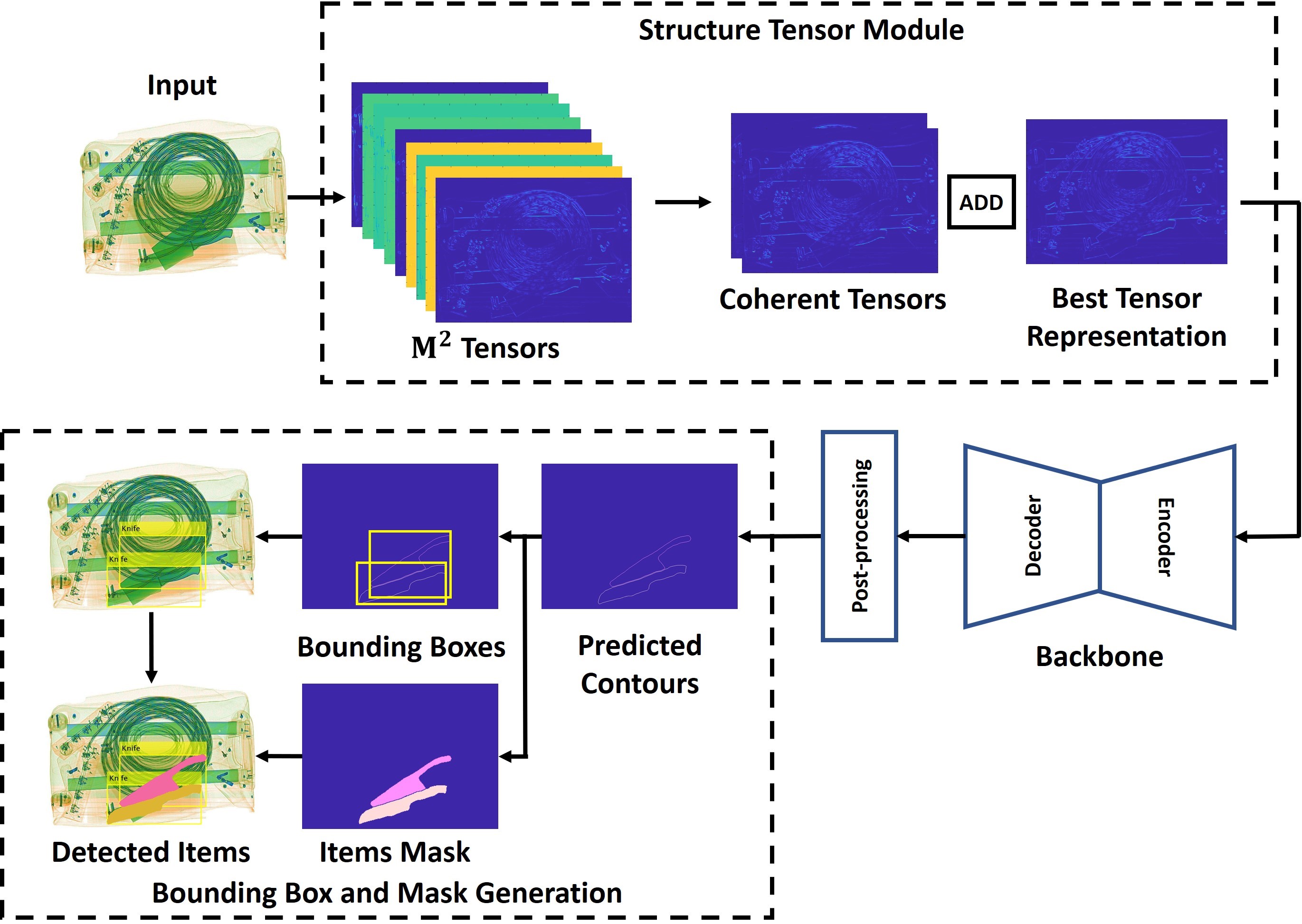}
    \caption{Block diagram of the proposed framework.}
    \label{fig:fig2}
\end{figure}

\subsection{Structure Tensor Module}
When the input X-ray scan is loaded into the proposed system, it is passed through the structure tensor module which highlights the transitions of each object (within the input scan) by summarizing the predominant orientations of its image gradients within the specified neighborhood of each pixel. However, the conventional structure tensor approach only computes the outer product of image gradients oriented at $x$ and $y$ direction, which limits the extraction of objects that are oriented only at these directions. To cater this, Hassan et al. \cite{hassan2019} recently proposed a modified structure tensor approach in which they compute the outer product of image gradients (dubbed tensor) oriented in any direction. Furthermore, instead of finding the strong orientations within the specified neighborhood of any pixel (as done in the conventional structure tensor), the modified structure tensor approach selects the coherent tensors according to their norm, such that they represent maximum transitions of each object within the candidate scan. But the modified structure tensor approach is still limited in differentiating between contours of the contraband items and normal baggage content. To overcome this, we present a novel trainable structure tensor scheme. Before discussing our approach, we first present both the original structure tensor and the modified structure tensor scheme for the sake of completeness.

\subsubsection{Structure Tensor:}
is a $2 \times 2$ matrix defined by the outer product of image gradients (oriented at $x$ and $y$ direction) within the specified neighborhood of each pixel. It summarizes the distribution of predominant orientations (for each object transition) within the associated image gradients and tells the degree to which these orientations are coherent. For the $i^{th}$ pixel in the input scan, the structure tensor $S(i)$ is defined as:

\begin{equation}
\resizebox{0.9\hsize}{!}{
$S(i)=$
$\begin{bmatrix}
  \sum_{j}\varphi_{(j)}(\nabla^{x}_{(i-j)})^2 & \sum_{j}\varphi_{(j)}(\nabla^{x}_{(i-j)}.\nabla^{y}_{(i-j)}) \\
  \sum_{j}\varphi_{(j)}(\nabla^{y}_{(i-j)}.\nabla^{x}_{(i-j)}) &
  \sum_{j}\varphi_{(j)}(\nabla^{y}_{(i-j)})^2
\end{bmatrix}$,}
\label{eq:eq1}
\end{equation}
\noindent where $\varphi$ is a Gaussian filter, $\nabla^{x}$ and $\nabla^{y}$ are the gradients oriented at $x$  and $y$ direction, respectively. Afterward, the degree of coherency or anisotropy is measured through:

\begin{equation}
c_d=\left(\frac{\lambda_1-\lambda_2}{\lambda_1+\lambda_2}\right)^2,
\end{equation}

\noindent where $c_d$ quantifies the degree of coherency, $\lambda_1$ and $\lambda_2$ are the eigenvalues of $S(i)$.

\subsubsection{Modified Structure Tensor:} The modified version of structure tensor can highlight the transitional patterns of the objects oriented in any direction by utilizing its respective image gradients \cite{hassan2019}. For the $M$ image gradients oriented at $M$ directions within the candidate scan, the modified structure tensor is defined as a $M \times M$ matrix such that:

\begin{equation}
\begin{bmatrix}
  \varphi * (\nabla^0 . \nabla^0) & \varphi * (\nabla^1 . \nabla^0) & \cdots & \varphi * (\nabla^{M-1} . \nabla^0) \\
  \varphi * (\nabla^0 . \nabla^1) & \varphi * (\nabla^1 . \nabla^1)  & \cdots & \varphi * (\nabla^{M-1} . \nabla^1) \\
  \vdots & \vdots & \ddots & \vdots \\
  \varphi * (\nabla^0 . \nabla^{M-1}) & \varphi * (\nabla^1 . \nabla^{M-1}) & \cdots & \varphi * (\nabla^{M-1} . \nabla^{M-1}) \\
\end{bmatrix}
\label{eq:eq2},
\end{equation}

\noindent where each tensor $\varphi * (\nabla^m . \nabla^n)$ is an outer product of image gradients (oriented at $m$ and $n$ direction) and the smoothing filter. Here, the gradient orientations ($\theta$) are computed as: $\vartheta = \frac{2 \pi \tau}{M}$ where $\tau$ varies from $0$ to $M-1$. Since, the multi-oriented block-structured tensor matrix (in Eq. \ref{eq:eq2}) is symmetrical, only the $\frac{M(M+1)}{2}$ (out of $M^2$) tensors are unique, and from these unique tensors, the most coherent ones (containing the maximum transitions of the baggage content) are selected according to their norm \cite{hassan2019}. Afterward, the selected tensors are added together to generate a single coherent representation of the object transitions as shown in Fig. \ref{fig:fig1} (B). 

\subsubsection{Proposed Trainable Structure Tensor:} The modified structure tensor approach (proposed in \cite{hassan2019}) can identify the transitional patterns of any object within the candidate scan by analyzing the predominant orientations of its image gradients (which leads towards the detection of concealed and cluttered contraband items \cite{hassan2019}). However, it cannot differentiate between the transitions of the contraband items and the normal baggage content. Therefore, to remove the noisy and irrelevant proposals it requires extensive screening efforts. Furthermore, CST framework \cite{hassan2019} uses an additional classifier to recognize the proposals of each contraband item within the candidate scan. Also, it has to be tuned for each dataset separately due to which it does not generalize well on multi-vendor X-ray scans \cite{hassan2019}. To address these limitations, we present a trainable structure tensor approach that employs an encoder-decoder backbone to extract and recognize contours of the highly cluttered, concealed, and overlapping contraband items, while suppressing the rest of the baggage content. The utilization of the encoder-decoder backbone eliminates the need for the additional classification network and it not only localizes each item through the bounding boxes but generates their masks as well. Therefore, unlike the object detection methods proposed in recent works \cite{miao2019sixray} \cite{opixray} \cite{hassan2019} for baggage threat detection, we propose a single-staged instance segmentation framework which, to the best of our knowledge, is the first framework in its category specifically designed to recognize the cluttered contraband items from the multi-vendor baggage X-ray scans.

\subsection{Bounding Box and Mask Generation}

After extracting and recognizing the contours of the contraband items, we first perform morphological post-processing to filter the false positives. Then, for each contraband item, we use its boundary to generate its bounding box (through the minimum bounding rectangle technique \cite{mbr}) and the mask (by filling the inner pixels with one's). 

\section{Experimental Setup} \label{sec:exp}

This section contains the details about the datasets, the implementation of the proposed framework, and the metrics which we used to validate the proposed framework.

\subsection{Datasets} 
The proposed framework has been validated on four publicly available grayscale and colored X-ray datasets. The detailed description of these datasets is presented below:

\subsubsection{GDXray} (first introduced in 2015 \cite{mery2015gdxray}) is the benchmark dataset for the non-destructive testing \cite{mery2015gdxray}. It contains 19,407 grayscale X-ray scans from the \textit{welds}, \textit{casting}, \textit{baggage}, \textit{nature}, and \textit{settings} categories. For the baggage threat recognition, the only relevant category is \textit{baggage} which contains 8,150 grayscale baggage X-ray scans containing suspicious items such as \textit{razors}, \textit{guns}, \textit{knives} and \textit{shuriken} along with their detailed ground truth markings. Moreover, we used 400 scans for the training purposes and the rest of the scans for the testing purposes as per the dataset standard \cite{mery2015gdxray}. 

\subsubsection{SIXray} (first introduced in 2019 \cite{miao2019sixray}) is another publicly available dataset containing 1,059,231 highly complex, and challenging colored baggage X-ray scans. Out of these 1,059,231 scans, 1,050,302 are negatives (that contains only the normal baggage content) whereas 8,929 scans are positive (containing the contraband items such as \textit{guns}, \textit{knives}, \textit{wrenches}, \textit{pliers}, \textit{scissors} and \textit{hammers}). Furthermore, the dataset contains detailed box-level markings of these items. To reflect the imbalanced nature of the positive scans, the dataset is arranged into three subsets namely SIXray10, SIXray100 and SIXray1000 \cite{miao2019sixray}. Moreover, each subset is partitioned into a ratio of 4 to 1 (i.e. 80\% of the scans were used for training while 20\% of the scans were used for the testing purposes as per the dataset standard \cite{miao2019sixray}). 

\subsubsection{OPIXray} (first introduced in 2020 \cite{opixray}) is a publicly available colored X-ray imagery dataset for the baggage threat detection \cite{opixray}. It contains a total of 8,885 X-ray scans from which 7,109 are arranged for the training purposes and the rest of 1,776 are dedicated for testing purposes \cite{opixray} (with the ratio of about 4 to 1). Furthermore, the datasets contains detailed box-level annotations of the five categories, namely, \textit{folding knives}, \textit{straight knives}, \textit{utility knives}, \textit{multi-tool knives}, and \textit{scissors}. In addition to this, the test scans within the OPIXray are categorized into three levels of occlusion \cite{opixray}.

\subsubsection{COMPASS-XP} (first introduced in 2019 \cite{compass}) is another publicly available dataset designed to validate the autonomous baggage threat recognition frameworks. The dataset contains matched photographic and X-ray imagery representing a single item in each scan. Unlike GDXray \cite{mery2015gdxray}, SIXray \cite{miao2019sixray}, and OPIXray \cite{opixray}, the COMPASS-XP dataset \cite{compass} is primarily designed for the image classification tasks where the ground truths are marked scan-wise indicating the presence of a normal or dangerous item within each scan. The total scans within COMPASS-XP dataset \cite{compass} are 11,568. From these scans, we used 9,254 for training purposes and the rest of 2,314 for testing purposes maintaining the ratio of 4 to 1.

\subsubsection{Combined Dataset:} Apart from validating the proposed framework on each of the four datasets separately, we combined them to validate the generalization capacity of the proposed framework against multiple scanner specifications. In a combined dataset, we used a total of 864,147 scans for training and the rest of 223,686 scans for the testing purposes.

\subsection{Implementation Details}
The proposed framework has been implemented using Python 3.7.4 with TensorFlow 2.2.0 and also the MATLAB R2020a on a machine having Intel Core i7-9750H@2.6 GHz processor and 32 GB RAM with a single NVIDIA RTX 2080 with cuDNN v7.5 and a CUDA Toolkit 10.1.243. The optimizer used during the training was ADADELTA \cite{adadelta} with a default learning and decay rate of 1.0 and 0.95, respectively. The source code is publicly available on GitHub\footnote{Source Code: \url{https://github.com/taimurhassan/TST}}.

\begin{table}[t]
    \centering
    \caption{Performance comparison of different encoder-decoder and fully convolutional networks in terms of DC and IoU for recognizing baggage threats. Bold indicates the best performance while the second-best performance is underlined.}
    \begin{tabular}{cccccc}
    \toprule
         Metric & Dataset & SegNet \cite{segnet} & PSPNet \cite{zhao2017pyramid} & UNet \cite{unet} & FCN-8 \cite{fcn8}\\\hline
         IoU & GDXray \cite{mery2015gdxray} & \textbf{0.7635} & 0.6953 & \underline{0.7564} & 0.6736 \\
         & SIXray \cite{miao2019sixray} & \textbf{0.6071} & 0.5341 & \underline{0.6013} & 0.4897 \\
         & OPIXray \cite{opixray} & \underline{0.6325} & 0.5689 & \textbf{0.6478} & 0.5018 \\
         & COMPASS-XP \cite{compass} & \textbf{0.5893} & 0.5016 & \underline{0.5743} & 0.4473 \\
         & Combined & \textbf{0.5314} & 0.4832 & \underline{0.5241} & 0.4103\\\hline
         DC & GDXray \cite{mery2015gdxray} & \textbf{0.8658} & 0.8202 & \underline{0.8613}	& 0.8049\\
        & SIXray \cite{miao2019sixray} & \textbf{0.7555} &	0.6963 & \underline{0.7510} & 0.6574\\
        & OPIXray \cite{opixray} & \underline{0.7748} & 0.7252 & \textbf{0.7862} & 0.6682\\
        & COMPASS-XP \cite{compass} & \textbf{0.7415} &	0.6680 & \underline{0.7295} & 0.6181\\
        & Combined & \textbf{0.6940} & 0.6515 &	\underline{0.6877} & 0.5818\\
    \toprule
    \end{tabular}
    \label{tab:tab1}
\end{table}

\begin{figure}[t]
    \centering
    \includegraphics[width=0.9\linewidth]{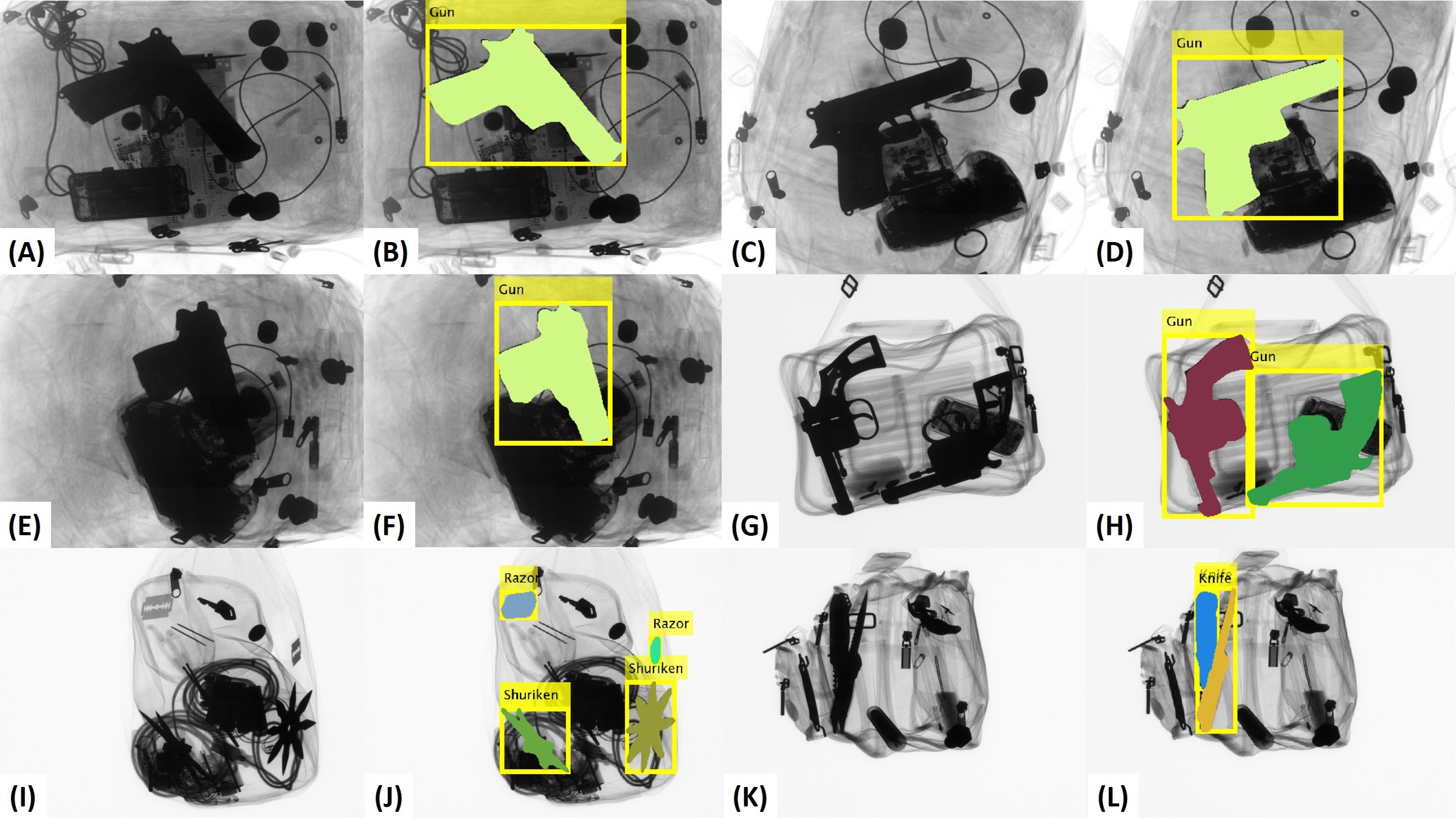}
    \caption{Visual examples showcasing the performance of the proposed framework on GDXray dataset.}
    \label{fig:fig3}
\end{figure}

\subsection{Evaluation Metrics}
The proposed framework has been evaluated using a variety of evaluation metrics as presented below:

\subsubsection{Dice Coefficient and Intersection-over-Union:} Dice Coefficient (DC) and Intersection-over-Union (IoU) measures the ability of the proposed framework that how accurately it has extracted the contraband items w.r.t the ground truths. DC is computed as $DC=\frac{2T_p}{2T_p + F_p + F_n}$ and IoU is computed as $IoU=\frac{T_p}{T_p + F_p + F_n}$, where $T_p$ denotes the true positives, $F_p$ denotes the false positives and $F_n$ denotes the false negatives. Moreover, we also computed the mean dice coefficient and mean IoU for each dataset by taking the average of DC and IoU scores, respectively for each contraband item category.

\subsubsection{Mean Average Precision} (mAP) is another metric that we used to validate the performance of the proposed framework for accurately detecting the prohibited items. The mAP scores in the proposed framework are calculated using the IoU threshold of 0.5.

\section{Results} \label{sec:results}

This section presents a thorough evaluation of the proposed framework on four publicly available datasets as well as on their combined representation. Furthermore, this section presents a detailed comparison of the proposed system with state-of-the-art frameworks. 

\subsection{Ablation Study}
We present an ablation study to determine 1) the optimal number of image gradients, their orientations, along with the number of coherent tensors which are to be selected within the structure tensor module to highlight the transitions of the contraband items, and 2) determining the optimal choice of encoder-decoder (or fully convolutional) backbone. 

\subsubsection{Number of Orientations and Coherent Tensors:}
Although including more image gradients (with more orientations) further reveals the transitional details of the baggage items within the candidate scan but it also makes the framework more vulnerable to noise and misclassifications. Similarly, considering more tensors (for generating the coherent representation) also affects the overall accuracy of the detection system \cite{hassan2019}. Although, the proposed framework is more robust against these issues as compared to the CST framework \cite{hassan2019}, mainly because of the removal of the irrelevant proposal screening process. But increasing the number of orientations and tensors do affect the ability of the encoder-decoder backbone to correctly recognize the boundaries of the contraband items. Therefore, after rigorous experimentation on each dataset, we selected the number of image gradients ($M$) to be 4 and the number of coherent tensors ($K$) to be 2, and these configurations have also been recommended in \cite{hassan2019}.

\subsubsection{Backbone Network:} The proposed framework employs an encoder-decoder or a fully convolutional network as a backbone to extract contours of the contraband item while simultaneously suppressing the rest of the baggage content. Here, we evaluated some of the popular architectures such as PSPNet \cite{zhao2017pyramid}, SegNet \cite{segnet}, UNet \cite{unet} and FCN \cite{fcn8}. Table \ref{tab:tab1} shows the comparison of these models on each dataset in terms of DC and IoU. We can observe that for the majority of the datasets, the best performance is achieved by the SegNet model \cite{segnet}, whereas the UNet \cite{unet} stood the second-best. Due to this, we prefer the use of SegNet \cite{segnet} as a backbone within the proposed framework. 

\begin{table}[t]
    \centering
    \caption{Performance comparison of the proposed framework on GDXray \cite{mery2015gdxray}, SIXray \cite{miao2019sixray} and OPIXray \cite{opixray} dataset in terms of mAP. Bold indicates the best performance while '-' indicates that the metric is not computed. CST and CHR frameworks are driven through ResNet-50 \cite{he2016deep}.}
    \begin{tabular}{cccccc}
    \toprule
        Dataset & Items & Proposed & CST \cite{hassan2019} & CHR \cite{miao2019sixray} & DOAM \cite{opixray} \\\hline 
        GDXray \cite{mery2015gdxray} & Knife & 0.9632 & \textbf{0.9945} & - & - \\
        & Gun & \textbf{0.9761} & 0.9101 & - & - \\
        & Razor & \textbf{0.9453} & 0.8826 & - & - \\
        & Shuriken & 0.9847 & \textbf{0.9917} & - & -  \\
        & \color{blue}{mAP} & \textbf{0.9672} & 0.9343 & - & -  \\\hline
        
        SIXray \cite{miao2019sixray} & Gun & 0.9734 & \textbf{0.9911} & 0.8640 & -  \\
        & Knife & \textbf{0.9681} & 0.9347 & 0.8536 & -  \\
        & Wrench & 0.9421 & \textbf{0.9915} & 0.6818 & -  \\
        & Scissor & 0.9348 & \textbf{0.9938} & 0.5923 & -  \\
        & Plier & \textbf{0.9573} & 0.9267 & 0.8261 & -  \\
        & Hammer & \textbf{0.9342} & 0.9189 & - & -  \\
        & \color{blue}{mAP} & 0.9516 & \textbf{0.9595} & 0.7635 & -  \\\hline
        
        OPIXray \cite{opixray} & Folding & 0.8024 & - & - & \textbf{0.8137}  \\
        & Straight & \textbf{0.5613} & - & - & 0.4150  \\
        & Scissor & 0.8934 & - & - & \textbf{0.9512}  \\
        & Multi-tool & 0.7802 & - & - & \textbf{0.8383} \\
        & Utility & \textbf{0.7289} & - & - & 0.6821  \\
        & \color{blue}{mAP} & \textbf{0.7532} & - & - & 0.7401  \\
    \bottomrule
    \end{tabular}
    \label{tab:tab2}
\end{table}

\subsection{Evaluations on GDXray Dataset}
First of all, we evaluated the proposed framework on the GDXray \cite{mery2015gdxray} dataset and also compared it with the state-of-the-art solutions as shown in Table \ref{tab:tab2}. Here, we can observe that the proposed framework obtained the mAP score of 0.9672 leading the second-best CST framework by 3.40\%. Although the proposed framework achieved the overall best performance in terms of mAP on the GDXray dataset, it lags from the CST framework by 3.14\% for extracting \textit{knives} and 0.705\% for extracting the \textit{shuriken}. Moreover, Fig. \ref{fig:fig3} shows the qualitative evaluations of the proposed framework where we can observe how robustly it has extracted the contraband items like \textit{guns}, \textit{shuriken} \textit{razors}, and \textit{knives}.

\begin{figure}[t]
    \centering
    \includegraphics[width=0.95\linewidth]{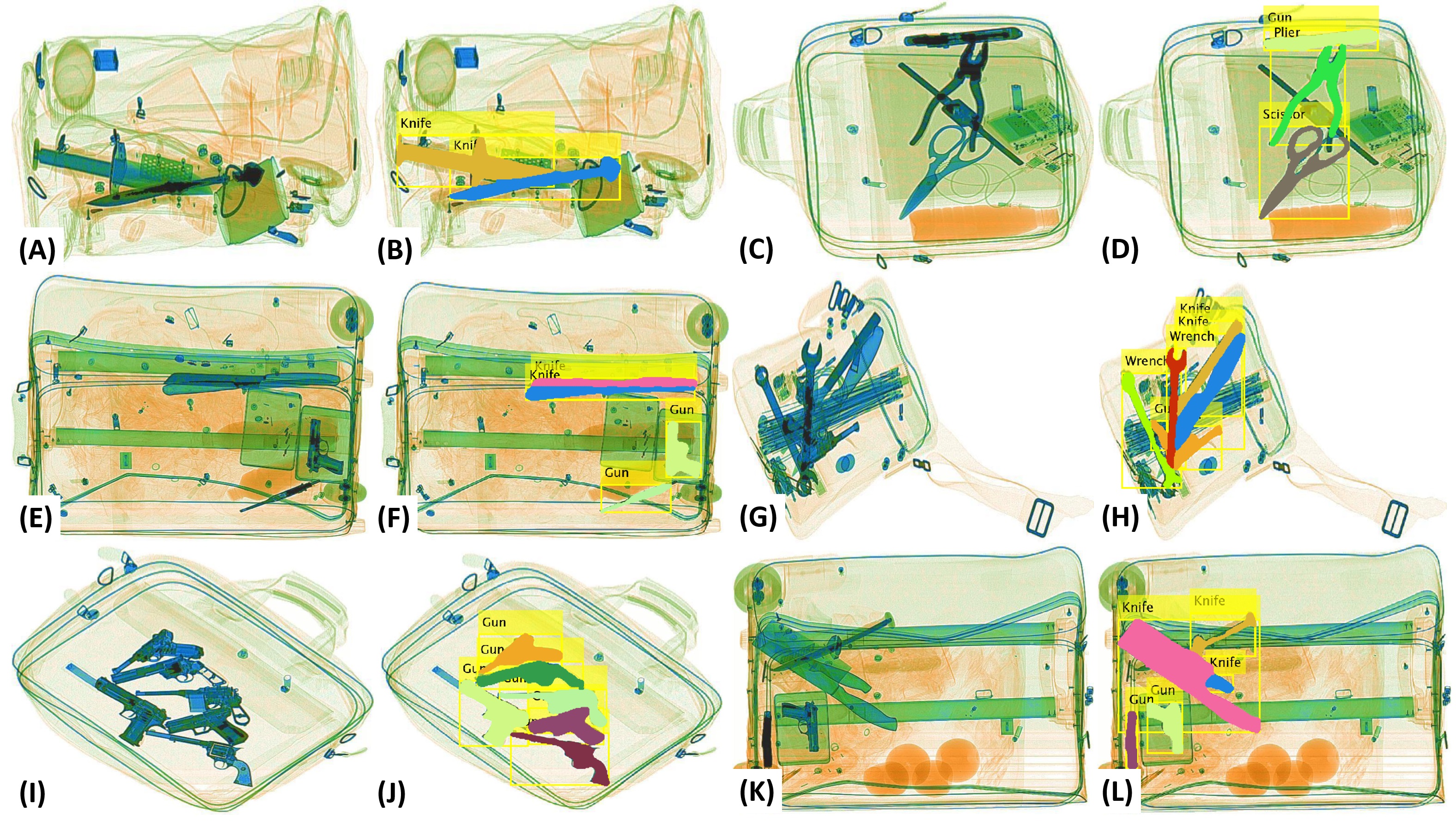}
    \caption{Visual examples showcasing the performance of the proposed framework on SIXray dataset.}
    \label{fig:fig4}
\end{figure}

\begin{figure}[htb]
    \centering
    \includegraphics[width=0.95\linewidth]{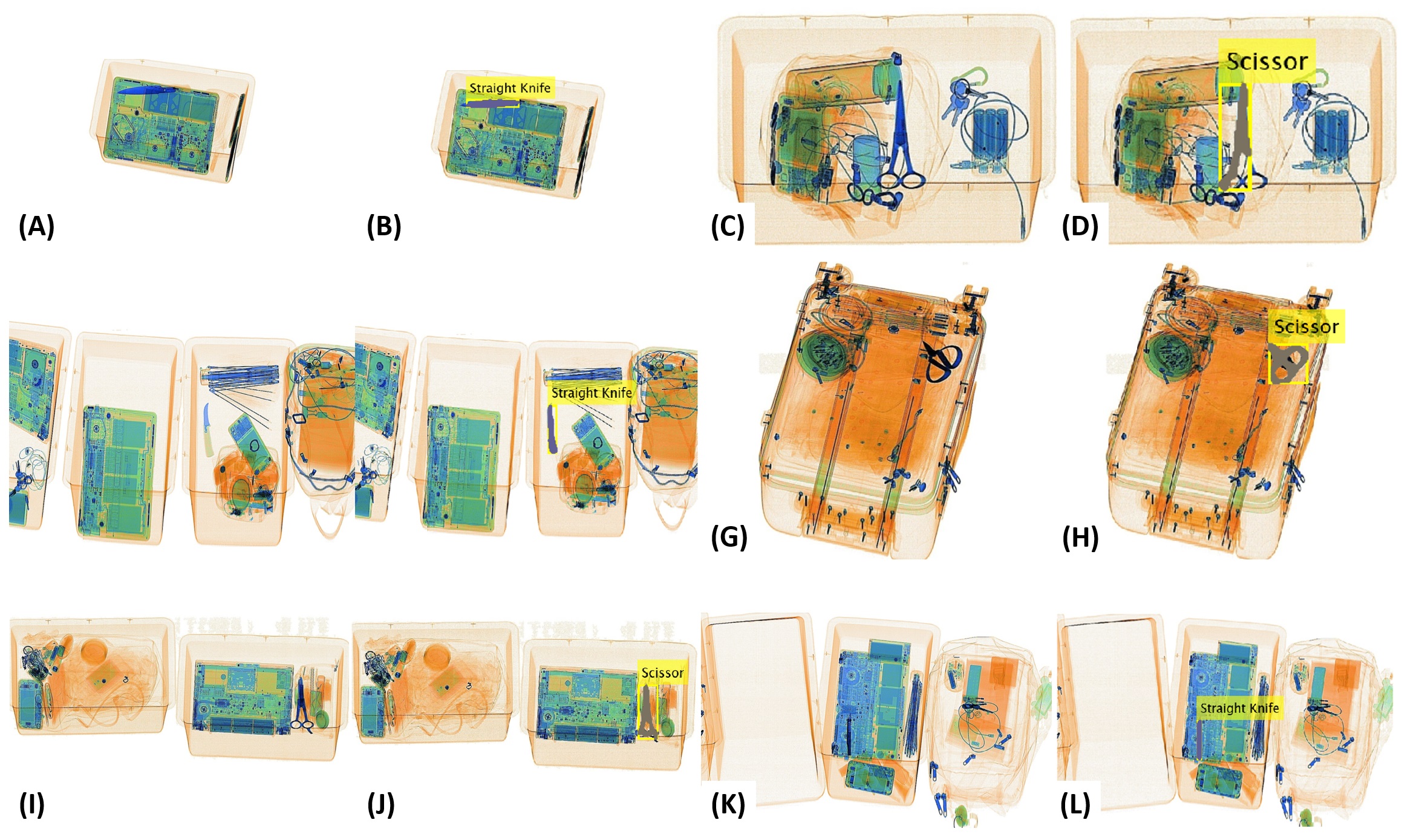}
    \caption{Visual examples showcasing the performance of the proposed framework on OPIXray dataset.}
    \label{fig:fig5}
\end{figure}

\begin{figure}[htb]
    \centering
    \includegraphics[width=0.95\linewidth]{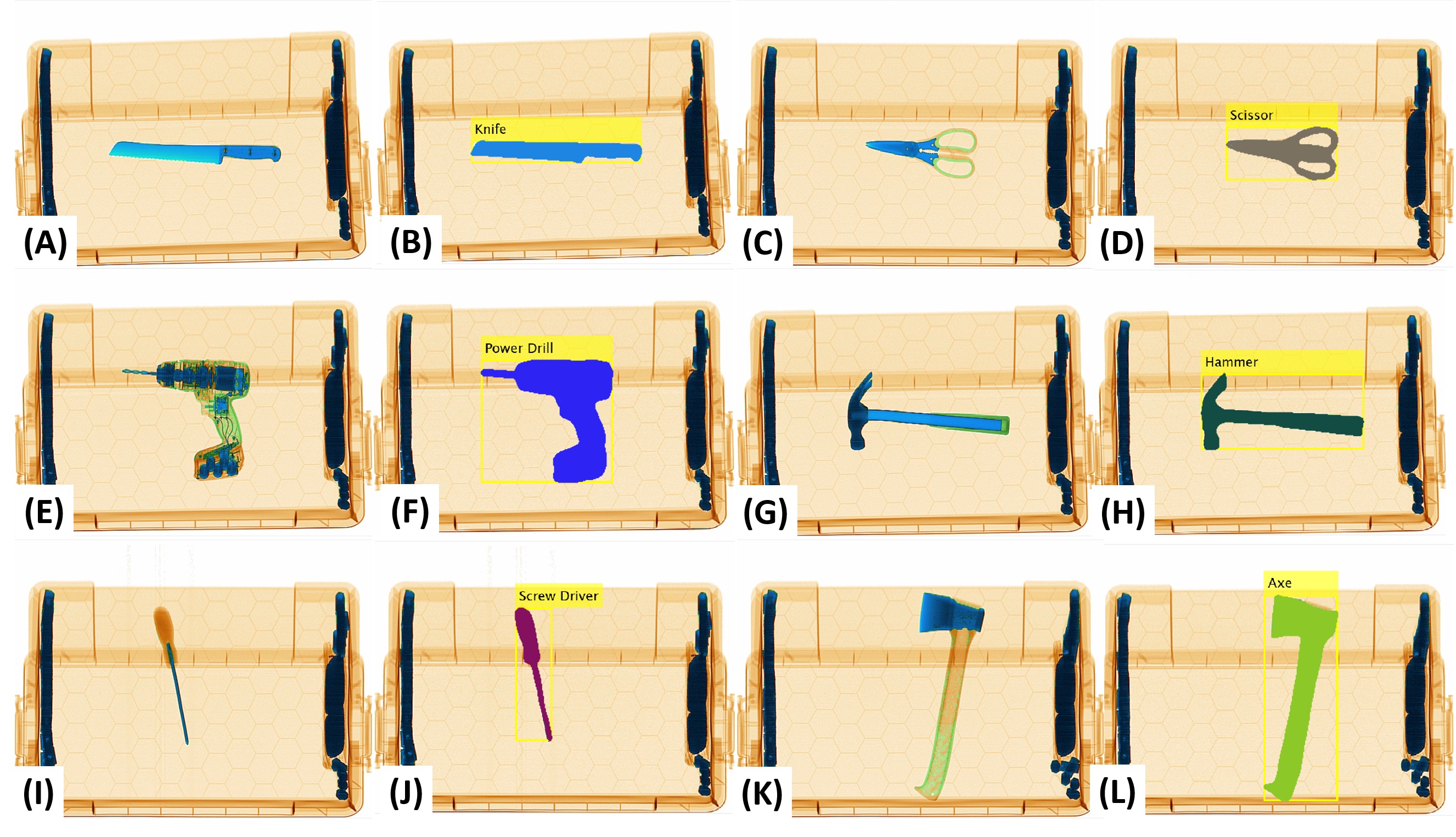}
    \caption{Visual examples showcasing the performance of the proposed framework on COMPASS-XP dataset.}
    \label{fig:fig6}
\end{figure}

\begin{figure}[htb]
    \centering
    \includegraphics[width=0.9\linewidth]{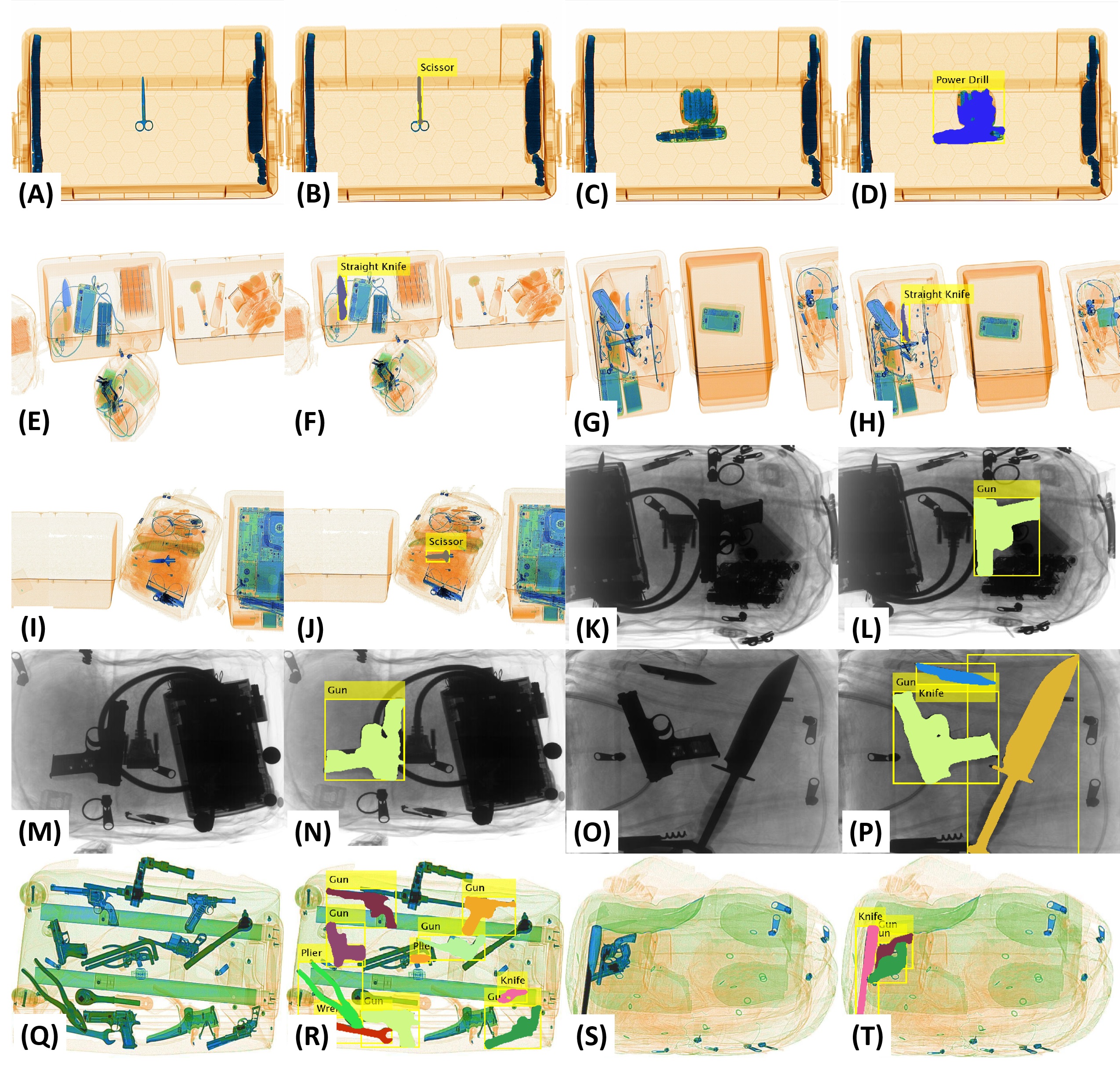}
    \caption{Visual examples showcasing the performance of the proposed framework on combined dataset.}
    \label{fig:fig7}
\end{figure}

\subsection{Evaluations on SIXray Dataset}
After evaluating the proposed framework on GDXray \cite{mery2015gdxray}, we tested it on the SIXray \cite{miao2019sixray}. SIXray, to the best of our knowledge, is one of the largest security inspection datasets containing highly complex baggage X-ray scans. In terms of mAP, the proposed framework stood second-best by achieving a score of 0.9516, lagging from the CST framework \cite{hassan2019} by only 0.823\% (please see Table \ref{tab:tab2}). However, it achieved a considerable edge over CST framework \cite{hassan2019} for extracting \textit{knife}, \textit{plier} and \textit{hammer} i.e. it leads CST \cite{hassan2019} by 3.45\% for detecting \textit{knives}, 3.19\% for recognizing \textit{pliers}, and 1.63\% for detecting \textit{hammers}. Apart from this, the performance of the proposed framework on each SIXray subset is shown in Table \ref{tab:tab3}. Here, we can observe that although the performance of the proposed framework is lagging behind CST framework \cite{hassan2019} on each subset, it's still outperforming CHR \cite{miao2019sixray} with a large margin. Moreover, on SIXray10, its lagging from CST framework \cite{hassan2019} by 0.342\% only. This is because the proposed framework possesses the ability to suppress the boundaries of normal baggage content. Even when it's trained on an imbalanced ratio of positive and negative scans, it shows a considerably good performance in recognizing the baggage threats. However, the CST framework \cite{hassan2019} stood first on each subset because it has been trained on the balanced set of object proposals in each subset. 

\begin{table}[htb]
    \centering
    \caption{Performance comparison of proposed framework with CST \cite{hassan2019} and CHR \cite{miao2019sixray} on each SIXray subset in terms of mAP. Bold indicates the best performance while the second-best performance is underlined. Here, CST \cite{hassan2019} and CHR \cite{miao2019sixray} are driven through ResNet-50 \cite{he2016deep}.}
    \begin{tabular}{cccc}
        \toprule
        Subset & Proposed & CST \cite{hassan2019} & CHR \cite{miao2019sixray} \\\hline
        SIXray10 & \underline{0.9601} & \textbf{0.9634} & 0.7794\\
        SIXray100 & \underline{0.8749} & \textbf{0.9318} & 0.5787\\
        SIXray1000 & \underline{0.7814} & \textbf{0.8903} & 0.3700\\
        \bottomrule
    \end{tabular}
    \label{tab:tab3}
\end{table}

\noindent Apart from this, the qualitative evaluation of the proposed framework on SIXray \cite{miao2019sixray} dataset is presented in Fig. \ref{fig:fig4}. 

\subsection{Evaluations on OPIXray Dataset}
The third dataset on which the proposed framework is evaluated is the OPIXray \cite{opixray}. OPIXray is the recently introduced dataset containing highly occluded color X-ray scans. From Table \ref{tab:tab2}, we can see that the proposed framework achieved the mAP score of 0.7532, outperforming DOAM \cite{opixray} by 1.73\%. From Table \ref{tab:tab2}, we can also observe that although DOAM has a considerable edge over the proposed framework for extracting \textit{folding knives}, \textit{multi-tool knives} and the \textit{scissors}. But since it lags from the proposed framework by 26.06\% for extracting \textit{straight knives} and by 6.42\% for extracting \textit{utility knives}, it stood the second-best. Apart from this, the qualitative evaluations of the proposed framework on the OPIXray dataset are shown in Fig. \ref{fig:fig5}. 

\subsection{Evaluations on COMPASS-XP Dataset}
The last dataset on which we evaluated the proposed framework is the COMPASS-XP dataset \cite{compass}. COMPASS-XP \cite{compass} is different than GDXray \cite{mery2015gdxray}, SIXray \cite{miao2019sixray} and OPIXray \cite{opixray} dataset as it contains the scans showcasing only a single item, and its primarily designed for evaluating the image classification frameworks. Nevertheless, we used this dataset for validating the performance of the proposed framework. The qualitative evaluations on COMPASS-XP \cite{compass} are shown in Fig. \ref{fig:fig6} where we can observe that the proposed framework is quite robust in picking different suspicious items. Apart from this, the proposed framework achieved the overall mAP score of 0.5842 on the COMPASS-XP dataset \cite{compass}.  Moreover, to the best of our knowledge, there is no literature available to date that utilizes the COMPASS-XP dataset \cite{compass} for validating the baggage threat detection framework. 

\subsection{Evaluations on Combined Dataset}
To evaluate the generalization capacity of the proposed framework on multi-vendor grayscale and colored X-ray scans, we combined all the four datasets and tested the proposed framework on the combined dataset. As observed in Fig. \ref{fig:fig7} that despite the large differences in the scan properties, the proposed framework has been able to accurately recognize the contraband items while generating good quality masks. Apart from this, the proposed framework achieved an overall mAP score of 0.4657 when evaluated on a diverse ranging 223,686 multi-vendor baggage X-ray scans. 

\noindent Here, we would also like to highlight that the proposed framework does get some false negatives (and false positives as well) while suppressing the irrelevant contours. Although, the false positives are handled through morphological post-processing. But, unfortunately, the proposed framework is somewhat limited to false negatives e.g. see the extracted \textit{scissor} in Fig. \ref{fig:fig5} (D), \textit{pliers} in Fig. \ref{fig:fig4} (D), \textit{scissor} in Fig. \ref{fig:fig7} (B), \textit{gun} in Fig. \ref{fig:fig7} (N), and \textit{pliers} in Fig. \ref{fig:fig7} (R). However, since the proposed framework is leveraged through pixel-wise recognition, we believe that this limitation is not drastic. Because even some pixels of the threatening items are classified as false negatives, the proposed framework does recognize the threatening items as a whole.

\section{Conclusion} \label{sec:conclusion}

This paper presents a single-staged instance segmentation framework capable of recognizing highly cluttered, concealed, and overlapping contraband items from the multi-vendor baggage X-ray scans. The proposed framework is based on a novel trainable structure tensor scheme that only highlights the transitional patterns of the contraband items, leading to their accurate detection. Furthermore, the proposed framework has been rigorously tested on four publicly available datasets where it outperformed state-of-the-art solutions in terms of mAP scores. In the future, the proposed detection framework can be extended to detect the contours of the 3D printed objects within the baggage X-ray scans. Furthermore, it can also be tested for object detection and instance segmentation on normal photographic imagery.  

%\bibliographystyle{lib/splncs}
%\bibliography{ref/egbib}

\begin{thebibliography}{10}

\bibitem{ackay2020}
Ak\c{c}ay, S., Breckon, T.:
\newblock Towards {A}utomatic {T}hreat {D}etection: {A} {S}urvey of {A}dvances of {D}eep
  {L}earning within {X}-ray {S}ecurity {I}maging.
\newblock preprint arXiv:2001.01293 (2020)

\bibitem{turcsany2013improving}
Turcsany, D., Mouton, A., Breckon, T. P.:
\newblock {I}mproving {F}eature-based {O}bject {R}ecognition for {X}-ray {B}aggage {S}ecurity {S}creening using {P}rimed {V}isual {W}ords
\newblock In: 2013 IEEE International Conference on Industrial Technology
  (ICIT), IEEE (2013)  1140--1145

\bibitem{bastan2015}
Bastan, M.:
\newblock Multi-view object detection in dual-energy {X}-ray images.
\newblock Machine Vision and Applications (2015)  1045–1060

\bibitem{heitz2010}
Heitz, G., Chechik, G.:
\newblock Object separation in x-ray image sets.
\newblock In: IEEE International Conference on Computer Vision and Pattern Recognition (CVPR). (2010)
  2093–2100

\bibitem{zhang2014}
Zhang, J., Zhang, L., Zhao, Z., Liu, Y., Gu, J., Li, Q., Zhang, D.:
\newblock Joint {S}hape and {T}exture {B}ased {X}-ray {C}argo {I}mage {C}lassification.
\newblock In: IEEE International Conference on Computer Vision and Pattern Recognition (CVPR) Workshops.
  (2014)  266–273

\bibitem{jaccard2014automated}
Jaccard, N., Rogers, T. W., Griffin, L. D.:
\newblock Automated detection of cars in transmission {X}-ray images of freight
  containers.
\newblock In: AVSS. (2014)  387--392

\bibitem{bastan2011}
Bastan, M., Yousefi, M. R., Breuel, T. M.:
\newblock Visual {W}ords on {B}aggage {X}-{R}ay {I}mages.
\newblock In: International Conference on Computer Analysis of Images and Patterns (CAIP).
  (2011)  360–368

\bibitem{kundegorski2016}
Kundegorski, M. E., Ak\c{c}ay, S., Devereux, M., Mouton, A., Breckon, T. P.:
\newblock On using feature descriptors as visual words for object detection
  within {X}-ray baggage security screening.
\newblock In: IEEE International Conference on Imaging for Crime Detection and Prevention (ICDP).
  (2016)

\bibitem{mery2016}
Mery, D., Svec, E., Arias, M.:
\newblock Object {R}ecognition in {B}aggage {I}nspection {U}sing {A}daptive {S}parse {R}epresentations of {X}-ray {I}mages
\newblock In: Pacific-Rim Symposium on Image and Video Technology. (2016)
  709–720

\bibitem{riffo2015automated}
Riffo, V., Mery, D.:
\newblock {A}utomated {D}etection of {T}hreat {O}bjects {U}sing {A}dapted {I}mplicit {S}hape {M}odel.
\newblock IEEE Transactions on Systems, Man, and Cybernetics: Systems
  46 (2015)  472--482

\bibitem{akcay2018using}
Ak\c{c}ay, S., Kundegorski, M. E., Willcocks, C. G., Breckon, T. P.:
\newblock {U}sing {D}eep {C}onvolutional {N}eural {N}etwork {A}rchitectures for {O}bject {C}lassification and {D}etection {W}ithin {X}-{R}ay {B}aggage {S}ecurity {I}magery
\newblock IEEE Transactions on Information Forensics and Security 13
  (2018)  2203--2215

\bibitem{jaccard2017}
Jaccard, N., Rogers, T. W., Morton, E. J., Griffin, L. D.:
\newblock Detection of concealed cars in complex cargo {X}-ray imagery using deep
  learning.
\newblock Journal of X-ray Science and Technology, (2017)  323–339

\bibitem{liu2018detection}
Liu, Z., Li, J., Shu, Y., Zhang, D.:
\newblock Detection and Recognition of Security Detection Object Based on Yolo9000
\newblock In: 2018 5th International Conference on Systems and Informatics
  (ICSAI), IEEE (2018)  278--282

\bibitem{Xu2018}
Xu, M., Zhang, H., Yang, J.:
\newblock Prohibited Item Detection in Airport X-Ray Security Images via Attention Mechanism Based CNN
\newblock In: Chinese Conference on Pattern Recognition and Computer Vision.
  (2018)  429–439

\bibitem{miao2019sixray}
Miao, C., Xie, L., Wan, F., Su, C., Liu, H., Jiao, J., Ye, Q.:
\newblock {S}{I}{X}ray: {A} {L}arge-scale {S}ecurity {I}nspection {X}-ray
  {B}enchmark for {P}rohibited {I}tem {D}iscovery in {O}verlapping {I}mages.
\newblock In: IEEE International Conference on Computer Vision and Pattern Recognition (CVPR). (2019)  2119--2128

\bibitem{gaus2019evaluating}
Gaus, Y. F. A., Bhowmik, N., Ak\c{c}ay, S., Breckon, T.:
\newblock Evaluating the Transferability and Adversarial Discrimination of Convolutional Neural Networks for Threat Object Detection and Classification within X-Ray Security Imagery
\newblock In: 18th IEEE International Conference On Machine Learning And Applications (ICMLA) (2019)

\bibitem{hassan2019}
Hassan, T., Bettayeb, M., Ak\c{c}ay, S., Khan, S., Bennamoun, M., Werghi, N.:
\newblock {D}etecting {P}rohibited {I}tems in {X}-ray {I}mages: {A} {C}ontour
  {P}roposal {L}earning {A}pproach.
\newblock In: 27th IEEE International Conference on Image Processing (ICIP).
  (2020)

\bibitem{gaus2019evaluation}
Gaus, Y. F. A., Bhowmik, N., Ak\c{c}ay, S., Guill{\'e}n-Garcia, P. M., Barker,
  J. W., Breckon, T. P.:
\newblock Evaluation of a Dual Convolutional Neural Network Architecture for Object-wise Anomaly Detection in Cluttered X-ray Security Imagery
\newblock In: 2019 International Joint Conference on Neural Networks (IJCNN).
  (2019)  1--8

\bibitem{an2019}
An, J.,  et~al.:
\newblock Semantic Segmentation for Prohibited Items in Baggage Inspection.
\newblock In: International Conference on Intelligence Science and Big Data Engineering. Visual
  Data Engineering (IScIDE). (2019)  495–505

\bibitem{akcay2018ganomaly}
Ak\c{c}ay, S., Atapour-Abarghouei, A., Breckon, T. P.:
\newblock G{A}{N}omaly: {S}emi-{S}upervised {A}nomaly {D}etection via
  {A}dversarial {T}raining.
\newblock In: Asian Conference on Computer Vision, Springer (2018)  622--637

\bibitem{samet2019}
Bhowmik, N., Gaus, Y. F. A., Ak\c{c}ay, S., Barker, J. W., Breckon, T. P.:
\newblock On the Impact of Object and Sub-component Level Segmentation Strategies for Supervised Anomaly Detection within X-ray Security Imagery
\newblock In: In Procedings of the International Conference on Machine Learning
  Applications (ICMLA). (2019)

\bibitem{griffin2019}
Griffin, L. D., Caldwell, M., Andrews, J. T. A., Bohler, H.:
\newblock “Unexpected Item in the Bagging Area”: Anomaly Detection in X-Ray Security Images
\newblock In: IEEE Transactions on Information Forensics and Security. (2019)  1539–1553

\bibitem{he2016deep}
He, K., Zhang, X., Ren, S., Sun, J.:
\newblock Deep {R}esidual {L}earning for {I}mage {R}ecognition.
\newblock In: IEEE Conference on Computer Vision and Pattern Recognition (CVPR). (2016)  770--778

\bibitem{mery2015gdxray}
Riffo, V., Lobel, H., Mery, D.:
\newblock GDXray: The Database of X-ray Images for Nondestructive Testing
\newblock Journal of Nondestructive Evaluation 34 (2015) 42

\bibitem{opixray}
Wei, Y., Tao, R., Wu, Z., Ma, Y., Zhang, L., Liu, X.:
\newblock {O}ccluded {P}rohibited {I}tems {D}etection: {A}n {X}-ray {S}ecurity
  {I}nspection {B}enchmark and {D}e-occlusion {A}ttention {M}odule (2020)

\bibitem{mbr}
Caldwell, D. R.:
\newblock {U}nlocking the {M}ysteries of the {B}ounding {B}ox.
\newblock Coordinates : Online Journal of the Map and Geography Round Table of
  the American Library Association. Series A (2005)

\bibitem{compass}
Griffin, L. D., Caldwell, M., Andrews, J. T. A.:
\newblock {COMPASS}-{XP} {D}ataset.
\newblock Computational Security Science Group, UCL (2019)

\bibitem{adadelta}
Zeiler, M. D.:
\newblock {A}{D}{A}{D}{E}{L}{T}{A}: {A}n {A}daptive {L}earning {R}ate {M}ethod.
\newblock arXiv:1212.5701 (2012)

\bibitem{segnet}
Badrinarayanan, V., Kendall, A., Cipolla, R.:
\newblock SegNet: A Deep Convolutional Encoder-Decoder Architecture for Image Segmentation
\newblock IEEE Transactions on Pattern Analysis and Machine Intelligence
  39 (2017)  2481--2495

\bibitem{zhao2017pyramid}
Zhao, H., Shi, J., Qi, X., Wang, X., Jia, J.:
\newblock Pyramid Scene Parsing Network.
\newblock In: IEEE International Conference on Computer Vision and Pattern Recognition (CVPR). (2017)  2881--2890

\bibitem{unet}
Ronneberger, O., Fischer, P., Brox, T.:
\newblock U-{N}et: {C}onvolutional {N}etworks for {B}iomedical {I}mage
  {S}egmentation.
\newblock International Conference on Medical Image Computing and Computer-Assisted Intervention (MICCAI). (2015)

\bibitem{fcn8}
Long, J., Shelhamer, E., Darrell, T.:
\newblock Fully Convolutional Networks for Semantic Segmentation.
\newblock In: IEEE International Conference on Computer Vision and Pattern Recognition (CVPR). (2015)  3431--3440

\end{thebibliography}

\end{document}